# Letters of the Alphabet:
# Discovering Natural Feature Sets

Ezana N. Beyenne
Northwestern University
ezananb@u.northwestern.edu

**Abstract -** Deep learning networks find intricate features in large datasets using the backpropagation algorithm.[1] This algorithm repeatedly adjusts the network connections' weights and examining the "hidden" nodes behavior between the input and output layer provides better insight into how neural networks create feature representations. Experiments that build on each other show that activity differences computed within a layer can guide learning.[2] A simple neural network is used, which includes a data set comprised of the alphabet letters, where each letter forms 81 input nodes comprised of 0 and 1s and a single hidden layer and an output layer. The first experiment aims to explain how the hidden layers in this simple neural network represent the input data's features. The second experiment attempts to reverse-engineer the neural network to find the alphabet's natural feature sets. As the network interprets features, we can understand how it derives the natural feature sets for a given data. This understanding is essential to delve deeper into deep generative models, such as Boltzmann machines. Deep generative models are a class of unsupervised deep learning algorithms. The primary function of deep generative models is to find the natural feature sets for a given data set.

## I. INTRODUCTION & PROBLEM STATEMENT

Deep network architectures find intricate features in large datasets using the backpropagation algorithm.[1] This algorithm repeatedly makes weight adjustments between the hidden nodes and output nodes to create the task domain's useful new features.[1] Examining hidden nodes' behavior provides better insight into how neural networks create feature representations within their hidden nodes and how they interpret these features. As the network interprets these features, we can understand how it derives the natural feature sets for a given data. This understanding is essential to delve deeper into deep generative models, such as Boltzmann machines. Deep generative models are a class of unsupervised deep learning algorithms. The primary function of deep generative models is to find the natural feature sets for a given data set.[3]

Many applications of deep learning networks use feedforward neural network architectures.[4] [4]These feedforward architectures learn to map a fixed-size input to a fixed-size output by going through a set of hidden layers of nodes that find intricate features in a non-linear way by using the backpropagation algorithm, and the categories become linearly separable by the output layer.[4] The first hidden layer of hidden nodes learns basic features, like lines and edges.[4]

Two experiments using a simple neural network show that activity differences computed within a layer can guide learning. Both include a data set comprised of the 26 letters of the English alphabet, where each letter forms 81 input nodes comprised of 0 and 1s and a single hidden layer and an output layer. In the first experiment, 26 letters are used as inputs to understand how the hidden layers in a simple-hidden layer network learn to represent features within a given input data. Each letter corresponds to a nine-by-nine element input grid and flattens to 81 input nodes. The hidden layer consists of six fully connected nodes, and the data of the output class are the 26 letters. The objective is to understand and interpret the hidden node activations as features without training a successful neural network. It identifies what the hidden nodes are reacting to for each different class of the input.

The second experiment builds upon the first and reverse-engineers the same neural network to find the natural feature sets. The output class is collapsed and runs from one letter for each output to a small number of classes. Each class represents a set of nine "similarly-shaped" letters, which form the natural feature sets. These "natural feature sets" reveal the sorts of hidden nodes in the neural network for each different class. This experiment adds noise at 10% to see if noise helps it generalize better. Table 1 below presents the classification of the "Natural Feature Sets".

| Natural Feature Sets | Letters of the alphabet |
|---|---|
| A | A, H |
| B | B, R, P |
| C | C, G |
| E | E, F, S |
| I | Z, T, I, J |
| K | Y, K, X |
| L | L, U |
| M | N, M |
| O | O, Q, D |
| V | V, W |

Table 1. Natural Feature Sets and similar letters

## II. LITERATURE REVIEW

Conventional machine learning techniques require careful engineering and domain expertise to design a feature extractor to transform raw data into suitable representations for the learning system to derive input patterns.[1] Representational learning is a set of algorithms that take raw data, which automatically derive the features needed for classification or detection tasks. [4]

Deep learning methods are representation learning algorithms with multiple layers of representation built-in.[5] These learning methods start with the raw data being input into a simple but non-linear layer that transforms the representation at one level into a representation at a higher, more abstract level.[6] The composition of enough such transformations can derive a very complex function.[6] For classification tasks, the higher layers of representation suppress irrelevant variations while amplifying aspects of the input that are important for discrimination.[6]

Deep learning algorithms are making significant advances in solving problems that have previously eluded the artificial intelligence community's best attempts because deriving natural feature sets from data happens without human engineers' intervention.[8] Deep learning has made it possible to find multiple layers of non-linear representations by backpropagating error derivatives through the feedforward neural network, as shown in Figure 1.[6]

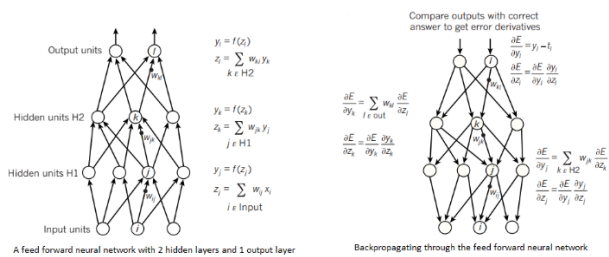

Figure 1. A multilayer neural network and backpropagation.[5]

Supervised deep learning algorithms are very productive when there is a massive amount of labeled training data.[6] The problem is that there is a lot more unlabeled data produced, and generative learning models are at the forefront to overcome the need for labeled data.[7] Generative learning algorithms are among the most promising approaches towards learning interesting features while attempting to understand this treasure trove of data.[7]

Boltzmann machines and Restricted Boltzmann machines belong to the unsupervised deep learning class. These algorithms use a generative learning methodology to make sense of unlabeled data.[8] Boltzmann machines are non-deterministic generative deep learning algorithms. They do not have output nodes, only hidden and visible nodes, as shown in Figure 2. All nodes connect, and this allows them to share information and self-generate subsequent data.[11] We only measure the visible nodes, and it takes raw data and can capture all the feature representations among the data.

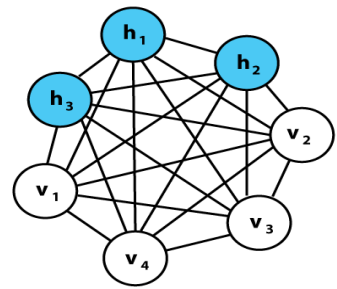

Figure 2. Boltzmann machines with hidden (h) and visible (v) nodes.[7]

Restricted Boltzmann machines (RBMs) are a class of Boltzmann machines with two-layered neural networks (one being the hidden layer and the other the visible layer) with generative capabilities, as shown in Figure 3.[8]

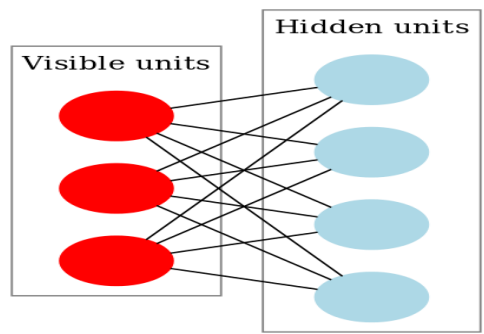

Figure 3. Restricted Boltzmann machine.[15]

RBMs have no connections between its hidden nodes, which makes learning much

---

more straightforward.[9] After a layer learns the data's hidden features; it becomes training data for other RBMs. When applied recursively, RBMs can learn a deep hierarchy of progressively more complicated features without needing labeled data.[7] RBMs as a feedforward neural network that can use backpropagation to train discriminatively. Multiple RBMs can be stacked to form a Deep Belief Network.[8] Backpropagation can discriminately fine-tune the weights in a feedforward neural network using a stack of RBMs. A stack of RBMs can work efficiently in much deeper networks while leading to better generalization.[6] Deep Belief Networks can efficiently train Boltzmann machines with millions of weights and many hidden layers.[6]

We can start to understand how generative deep learning models make sense of raw unlabeled data by considering how the hidden layers in a simple-hidden layer network learn to represent features to determine the input data's natural feature sets. Additionally, neural networks have usually been considered a black box with input going in; learning happens in the hidden layers to find a representation of input data and then have some output expectations.[9] By the end of these experiments, the intent is that neural networks will become more transparent.

## II. DATA

This analysis comprises of the 26 letters of the English alphabet. Each letter comprises 81 input nodes, and it corresponds to having a nine-by-nine grid with the filled-in blocks working together to create an image of the letter. The 81-node input layer is a one-dimensional array (e.g., an 81 by one format). The 81-node input layer is seen in a nine-by-nine grid so that it is easier to interpret visually, as shown in Figure 4. Figure 4 gives an example of the letters A, B, Y, and Z. All the letters display a level of consistency in that they take up the entire nine by nine grid in their representations. For the network to learn features, consistency is essential.

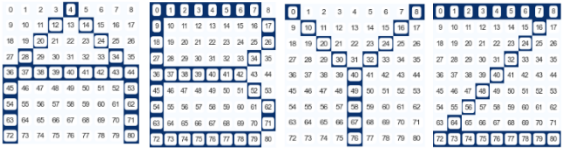

Figure 4. 9 by 9 Letter representations of the letters A, B, Y, and Z

The output classes differentiate the first and second experiments. In the first experiment, the output class consists of the 26 letters of the alphabet with one letter for each output. In the second experiment, the output class is collapsed and runs from one letter for each output to a smaller number of classes. Each class represents a set of nine "similarly-shaped" letters, which form the natural feature sets, as shown in Table 1. These "natural feature set classes" reveal the sorts of hidden nodes in the neural network for each different class.

## IV. RESEARCH DESIGN AND MODELING METHODS

The neural network comprises 81 input nodes comprised of a single fully connected hidden layer with six hidden nodes. The hidden node activations experiment has an output layer of size 26 to represent the possible letter outputs. The natural feature sets experiment has an output layer of size 9 to represent a set of nine "similarly-shaped" letters which form the natural feature sets. Noise is added to the natural feature sets' input to see if it improves generalization. A sigmoid transfer function with alpha set to one calculates the hidden layers' output. For the learning method, I used the backpropagation algorithm to train the neural network with a learning rate (eta) set to 0.5 that iterates over 5,000 epochs. The resulting total sum of squared errors reaches the set epsilon threshold of 0.01. I initially started with ten nodes for the hidden layer, then compared it with an eight-layer node, and finally chose six nodes for the hidden layer. The neural network with six nodes had better generalization.

## V. HIDDEN NODE ACTIVATIONS EXPERIMENT

### Results

A heatmap visualizes the weighted matrix of the hidden node activations. The hidden nodes perform simple pattern recognition and abstract away different features from the inputs and biases. By looking at the heatmaps, we can determine which combinations of the input data contributed to the hidden node activation values and therefore see which features each node picks. The red areas in the heatmap indicate parts of the matrix with an increased chance of the input causing an activation, whereas the yellow areas indicate a decreased chance. Figure 5 shows a nine-by-nine weighted heatmap for hidden node activations 1 through 6. Figure 5 shows the extracted representations of the features for each of the letters.

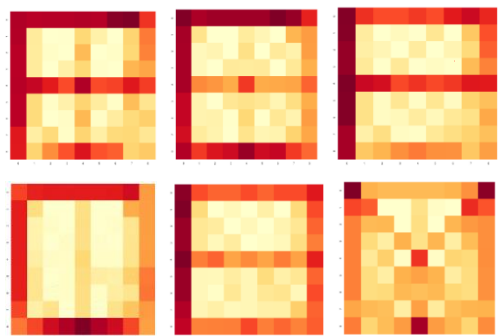

Figure 5. Hidden node activations 1, 2, 3, 4, 5, 6 in a nine by nine pixel grid

### Analysis and interpretation

Table 2 shows a summary of the feature extracted from Figure 5's heatmap. The table shows the letters that activate each node and the extracted features for the six hidden nodes.

| Node | Letters | Features extracted |
|------|---------|--------------------|
| 1 | P, F, E, R, B, I, U, T | Vertical bar along the left and somewhat middle, horizontal bar through the top, middle, somewhat the bottom, right vertical bar through the top half. Fails to turn the diagonals. Strongest for P, F, E. |
| 2 | O, P, E, F, D, B, G, L, C, S, Z, I, Q, R, Y | Vertical bar on the left and right, the top, left, bottom, somewhat middle horizontal bar, middle pixel, shows a weak turn on the diagonals for a faded Y. Strong showing for O, E, P, G. |
| 3 | P, R, | Vertical bar along the left, horizontal bar through the top, middle and somewhat the bottom, right vertical bar through the top half. |
| 4 | C, D, G, O, I, Q, Q, U | Vertical left and somewhat right bars, horizontal top and bottom bars, middle vertical bar although this is not as strong. |
| 5 | Every letter but I, T, X, Y | Vertical bars on the left and right, horizontal bars on the top, bottom and middle. Shows a faint diagonal. |
| 6 | T, V, W, X, M, N, Z | Top left and right pixels, along with bottom middle pixel, left and right and somewhat middle vertical bars, somewhat top horizontal bar. |

Table 2. Analysis of feature extractions from each hidden node in a 6-hidden node network

As we have seen, nodes 2 and 5 are activated strongly by most of the letters. The strong activation by most letters in the nodes mentioned above indicates that too much freedom might exist within the network for the hidden nodes to drop out. The next experiment groups these letters into similarly shaped letters that form its natural feature set.

## VI. NATURAL FEATURE SETS EXPERIMENT

### *Results*

The algorithm generated a nine-by-nine diagram for each hidden node for each unique letter belonging to the natural feature set. The darker the color in the heatmap, the more likely the hidden node identifies as the unique letter belonging to the natural feature set. We first generate the nine-by-nine diagrams for each hidden node with no noise and then rerun it with the addition of 10% noise. The intent is to see if noise helps each hidden node activate better for the feature set's unique letters. The top hidden row of nodes in the figures have no noise associated with them, whereas the bottom row has a 10% noise.

Figure 6 illustrates the natural feature set for the letter 'A', along with the six hidden nodes showing activations for its unique letters' A' and 'H'. Appendix A includes the nine-by-nine heatmaps with and without the noise of the nine natural feature sets. An in-depth analysis and interpretation appear in the next section using Appendix A's heatmaps.

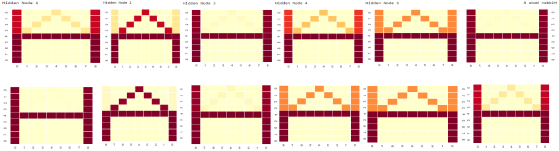

Figure 6. Natural Feature Set A and unique letters: ['A', 'H']

### *Analysis and interpretation*

In Figure 6, the natural feature set 'A' shows the hidden node activation of unique letters 'A' and 'H' where nodes 1 and 6 seem to have flipped. Overall, the heatmaps without noise were better at generalizing the unique letters. For example, node 2 and node 4 show a deeper color

---

[10] Elgendy, Mohamed. Deep Learning for Vision Systems. (Manning Press, October 2020).252-253

red for 'H' and 'A'. In appendix A, the hidden node activations for the natural feature set 'B' and its unique letters 'B', 'R', 'P' with and without noise appear to have similar outcomes. Hidden nodes 3 and 6 perform slightly better on 'B' and 'R', but hidden node 1 without noise performs better with 'B'.

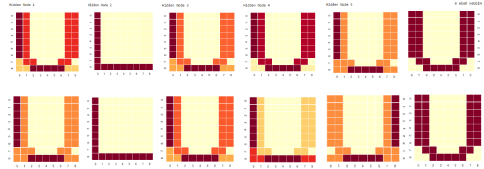

Figure 7. Natural Feature Set L and unique letters: ['U','L']

The heat maps shown in Figure 8 for natural set V with the unique letters 'V' and 'W' illustrates that hidden 4 without noise and hidden node 1 with noise are identical. For most of the hidden node activations for the natural feature set V, there are similarities between those with and without noise, with the sole exception being hidden node 4 with noise. This node recognizes only 'W'.

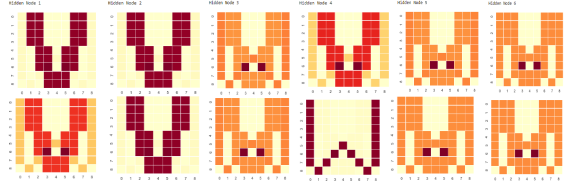

Figure 8. Natural Feature Set V and unique letters: ['V','W']

## VII. DESIGN AND IMPLEMENTATION CONSIDERATIONS

A neural network learns features by going through each fully connected hidden layer, which learns the features in an increased complexity level.[10] The first hidden layer with only six nodes, which we used, was only expected to learn low-level features.[11] Determining the number of hidden nodes was a matter of trial and error. The input was a letter of the alphabet; going with a six-hidden node layer led to getting the model trained faster, without much difference from the ten-hidden node layer. Instead of spending a lot of time determining the number of hidden nodes, I might consider a fuzzy logic approach.[12] As an alternative to the fuzzy logic approach, using the Separability-correlation measure to determine the number of hidden layers and neurons of the deep network.[13]

## VIII. CONCLUSION

The objective was first to understand how the hidden node activations in a neural network create feature representations for the English letters' input data. The first experiment's heatmaps showed the letters activated strongly for each hidden node as lines and edges. In the next experiment, the 26 letters of the alphabet were grouped into 9 "similarly shaped" classes called the data's natural feature sets. The heatmaps for each of the natural feature sets displayed the unique letters associated with them. I was expecting the addition of 10% noise to help the hidden nodes generalize better, but that did not turn out to be the case. Deep generative models build upon the lower layers, generating simple features and finding the natural feature sets for a given data set. The two experiments give a step-by-step blueprint into how natural feature sets are derived.

## IX. FUTURE DIRECTIONS

Deciphering the hidden node activations and determining how to visualize the information without a lot of explanation took a lot of time. The visualizations using heatmaps for the hidden nodes in the first experiment provided a clear explanation. Using the same heatmap of the hidden nodes for all the natural feature sets did not convey the message. I had to find a way to show the hidden node activations' heatmap for each of the nine natural feature sets. I would have liked to spend more time formatting them so that each of the six-hidden node heatmaps showed up on one line without taking up space. Trying some of the algorithms to automatically determine the ideal number of hidden nodes would help speed up development time. Tuning the number of hidden nodes is still considered a difficult task even for expert designers.[11]

---

[11] Saeed Pirmoradi et al., "The Self-Organizing Restricted Boltzmann Machine for Deep Representation with the Application on Classification Problems," Expert Systems with Applications, Volume 149, 1 July 2020, 113286 (Pergamon, February 13, 2020), https://www.sciencedirect.com/science/article/abs/pii/S0957417420301111.

[12] M. -. Chow, R. N. Sharpe and J. C. Hung, "On the application and design of artificial neural networks for motor fault detection. II," in *IEEE Transactions on Industrial Electronics*, vol. 40, no. 2, pp. 189-196, April 1993, doi: 10.1109/41.222640.

APPENDIX A – NATURAL FEATURE SETS

<u>Natural Feature Set</u> - **A**
Unique letters in this Natural Feature Set: ['A', 'H']
    ***a.*** *Hidden node activations with no noise*

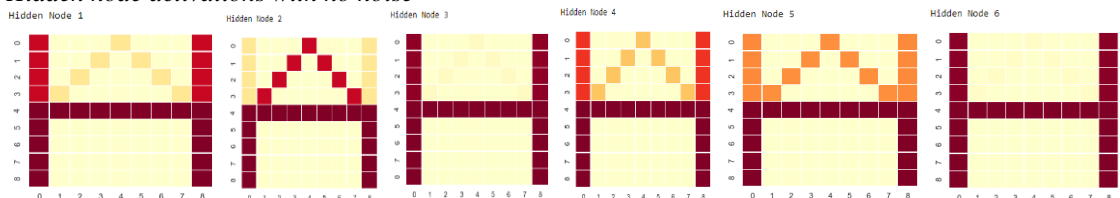

    ***b.*** *Hidden node activations with 10% noise.*

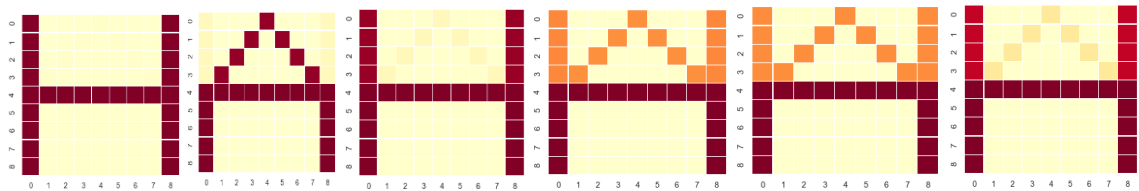

<u>Natural Feature Set</u> - **B**
Unique letters in this Natural Feature Set: ['B', 'R','P']
    **a.** Hidden node activations with no noise

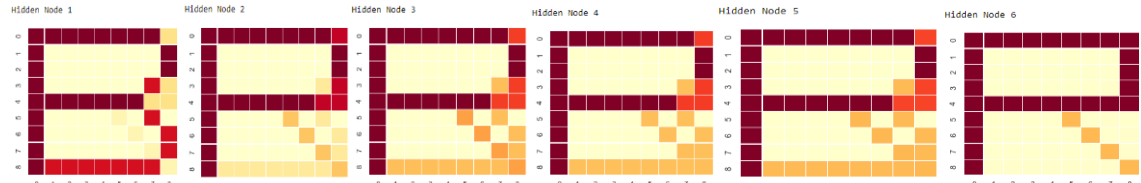

    **b.** Hidden node activations with 10% noise.

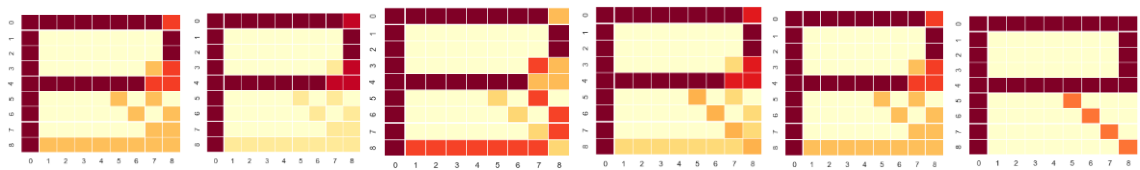

<u>Natural Feature Set</u> - **C**
Unique letters in this Natural Feature Set: ['C', 'G']
    **a.** Hidden node activations with no noise

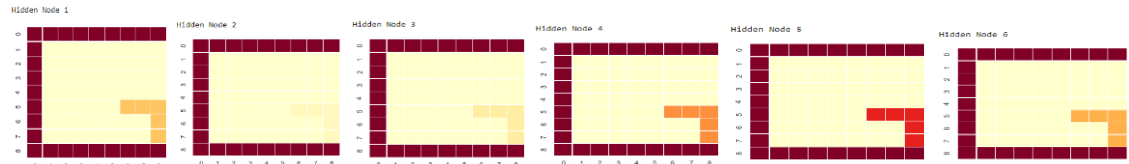

    **b.** Hidden node activations with 10% noise.

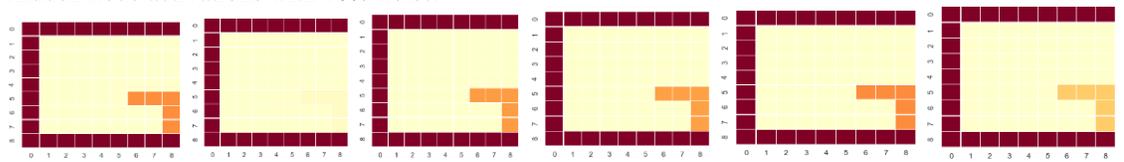

<u>Natural Feature Set</u> - **E**
Unique letters in this Natural Feature Set: ['E', 'F','S']

**a.** Hidden node activations with no noise

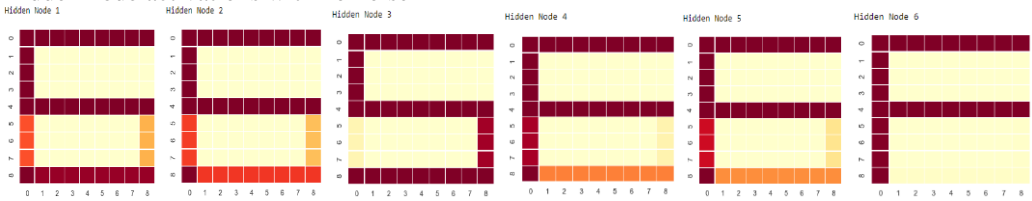

**b.** Hidden node activations with 10% noise.

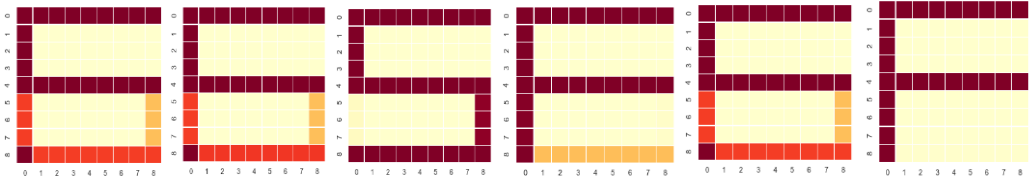

<u>Natural Feature Set</u> - **I**
Unique letters in this Natural Feature Set: ['J', 'I', 'Z', 'T']

    **a.** Hidden node activations with no noise

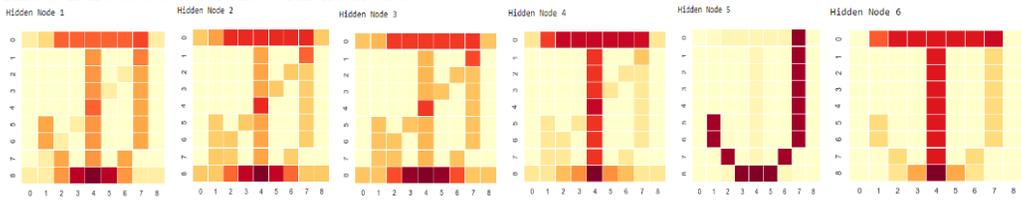

    **b.** Hidden node activations with 10% noise.

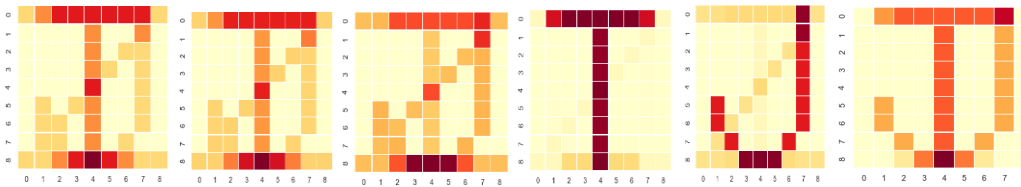

<u>Natural Feature Set</u> - **K**
Unique letters in this Natural Feature Set: ['Y', 'K', 'X']

    **a.** Hidden node activations with no noise

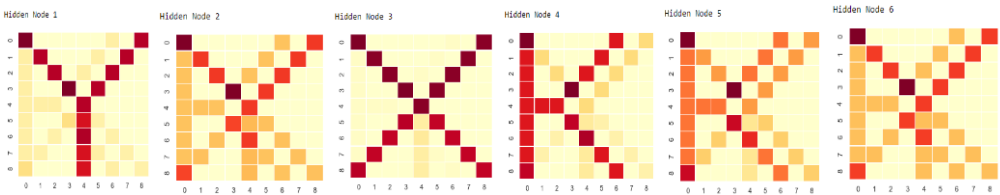

    **b.** Hidden node activations with 10% noise.

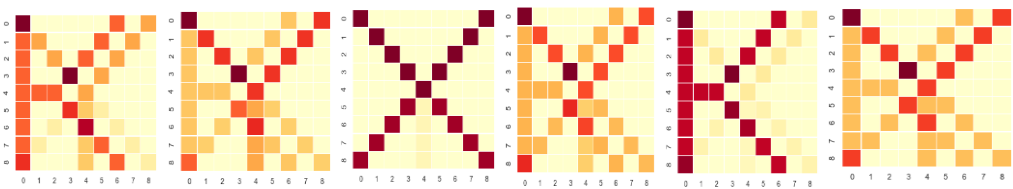

## Natural Feature Set - **L**

Unique letters in this Natural Feature Set: ['U','L']

    **a.** Hidden node activations with no noise

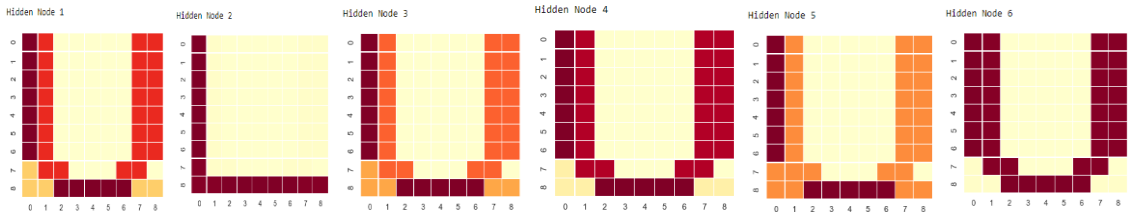

    **b.** Hidden node activations with 10% noise.

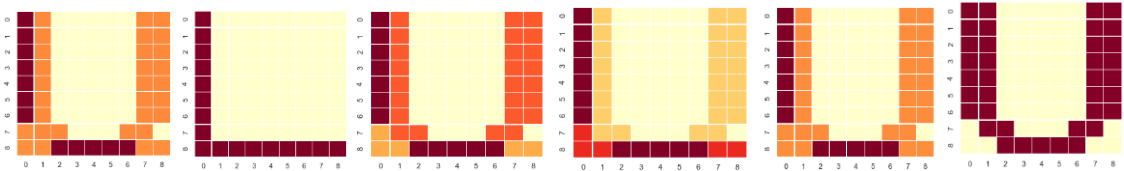

## Natural Feature Set - **M**

Unique letters in this Natural Feature Set: ['M','N']

    **a.** Hidden node activations with no noise

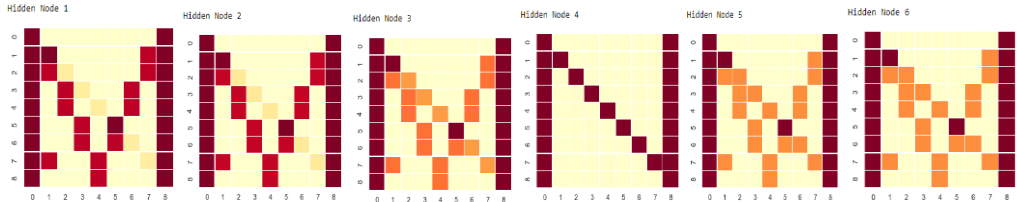

    **b.** Hidden node activations with 10% noise.

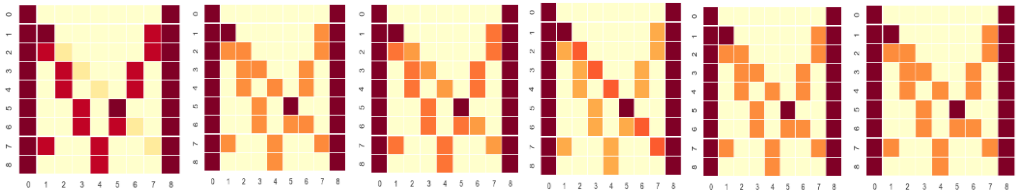

## Natural Feature Set - **O**

Unique letters in this Natural Feature Set: ['O','D','Q']

    **a.** Hidden node activations with no noise

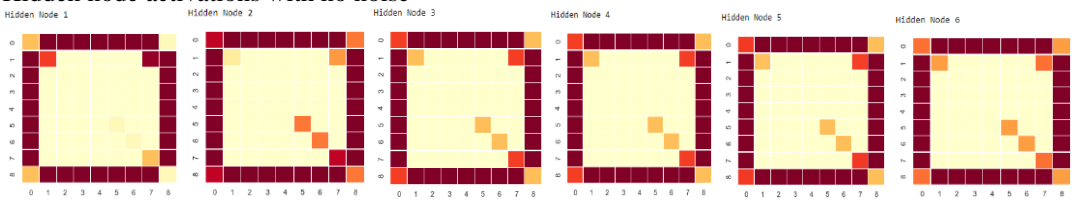

    **b.** Hidden node activations with 10% noise.

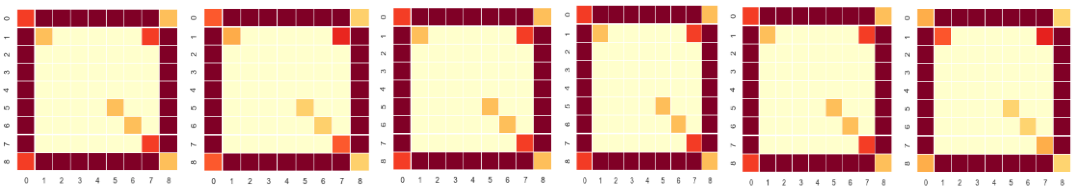

<u>Natural Feature Set</u> - **V**

Unique letters in this Natural Feature Set: ['V','W']

    **a.** Hidden node activations with no noise

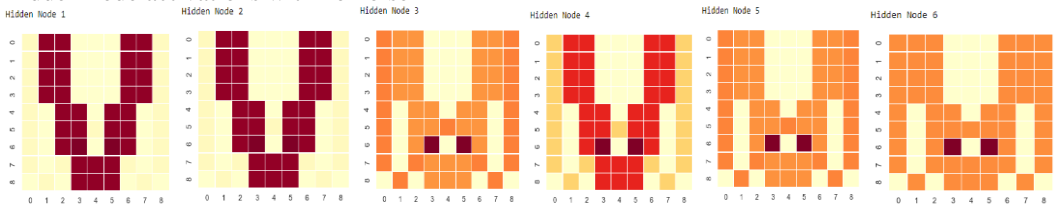

    **b.** Hidden node activations with 10% noise.

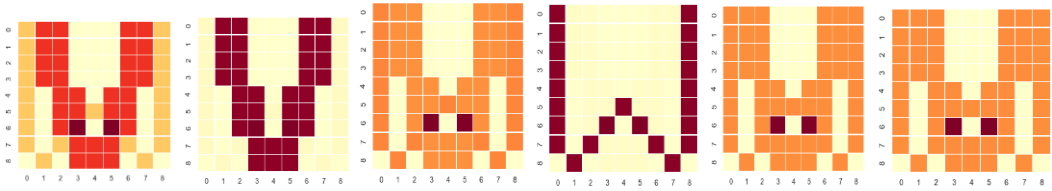